\documentclass[twocolumn,twoside]{IEEEtran}
\usepackage{graphicx}
\usepackage{cite}
\usepackage{url}
\usepackage{siunitx}
\usepackage{authblk}
\usepackage{balance}
\usepackage{amsmath}
\usepackage{mathtools} 
\usepackage{algorithm2e}
\usepackage{multirow}
\usepackage{float}
\usepackage{cite}
\usepackage{notoccite}
\usepackage{url}
\usepackage{amsmath}
\usepackage{color,soul}
\usepackage{lipsum} 
\usepackage{longtable}
\usepackage{graphicx}
\usepackage{atbegshi}
\usepackage{tabto}
\floatstyle{plaintop}
\restylefloat{table}

\usepackage{textcomp}
\usepackage{gensymb,siunitx}

\usepackage{booktabs,tabulary}
\usepackage{siunitx}
\usepackage{threeparttable}
\usepackage{subcaption}
\begin{document}


\bstctlcite{IEEEexample:BSTcontrol}

\author{Dima Kilani, Baker Mohammad, Yasmin Halawani, Mohammed F. Tolba and Hani Saleh}
\affil{System-on-Chip Center (SoCC), Khalifa University, Abu Dhabi, UAE}

\thispagestyle{empty}
\title{C3PU: Cross-Coupling Capacitor Processing Unit Using Analog-Mixed Signal In-Memory Computing for AI Inference} 
\maketitle
\thispagestyle{empty}
\pagestyle{empty}

\begin{abstract}

This paper presents a novel cross-coupling capacitor processing unit (C3PU) that supports analog-mixed signal in-memory computing to perform multiply-and-accumulate (MAC) operations. The C3PU consists of a capacitive unit, a CMOS transistor, and a voltage-to-time converter (VTC). The capacitive unit serves as a computational element that holds the multiplier operand and performs multiplication once the multiplicand is applied at the terminal. The multiplicand is the input voltage that is converted to a pulse width signal using a low power VTC.  The transistor transfers this multiplication where a voltage level is generated. A demonstrator of 5$\times$4 C3PU array that is capable of implementing 4 MAC units is presented. The design has been verified using Monte Carlo simulation in 65 nm technology. The 5$\times$4 C3PU consumed energy of 66.4 fJ/MAC at 0.3 V voltage supply with an error of 5.7\%. The proposed unit achieves lower energy and occupies a smaller area by 3.4$\times$ and 3.6$\times$, respectively, with similar error value when compared to a digital-based 8$\times$4-bit fixed point MAC unit. The C3PU has been utilized through an iris flower classification utilizing an artificial neural network which achieved a 90\% classification accuracy compared to ideal accuracy of 96.67\% using MATLAB.

\end{abstract}




\section{Introduction}

Multiply-and-accumulate (MAC) units are essential building blocks for digital processing units that are used in a multitude of applications, including artificial intelligence (AI) for edge devices, signal/image processing, convolution, and filtering \cite{masadeh2019input}. 
Recently, research has been focused on AI applications to address complex machine learning problems such as image/speech recognition and language translation \cite{ambrogio2018equivalent}. Deep neural networks (DNNs) are widely utilized in such applications since it can achieve high accuracy \cite{guan2019recursive}. However, DNN algorithms are computationally intensive, with large data sets that require high memory bandwidth. This results in memory access bottlenecks that introduce considerable energy and performance challenges. The memory access energy is 1-3 orders of magnitude higher than the compute energy \cite{han2016eie}. 
However, DNNs are approximate in nature, and many AI applications can tolerate lower accuracy \cite{wang2019deep}. This opens the opportunity for potential tradeoffs between energy efficiency, accuracy, and latency. 

One direction to reduce the need for explicit memory access is to utilize in-memory computing (IMC) architectures. It has significant advantages in energy efficiency and throughput compared to traditional computing that is based on von Neumann architecture \cite{jiang2019c3sram} \cite{halawani2019reram} \cite{halawani2018memristor}. Both digital and analog approaches for IMC have been reported in the literature to develop an artificial neural network (ANN). One key component in the ANN is the synaptic memory used for the network's weight storage. Several weight storage elements are reported and classified in the literature: $1)$ traditional volatile memory: SRAM \cite{si2019twin} and DRAM \cite{gao2019computedram}, $2)$ non-volatile memory (NVM): CMOS-based flash memory \cite{du2018analog}, and emerging NVM technology such as Resistive RAM (RRAM) \cite{hu2018memristor} and $3)$ analog-mixed signal circuits using capacitors and transistors \cite{kim2017analog}. Both SRAM and DRAM are limited to high power devices that are not suitable for duty-cycled edge devices. The flash memory traps the weight charges in the floating gate, which is electrically isolated from the control gate. On the other hand, the RRAM devices such as memristors store the weight as a conductance value. However, memristors suffer from low endurance and sneak path issues which may result in a state disturbance \cite{demme2015increasing}. Capacitors and transistors structures have been demonstrated by IBM as an analog memory to store the weights as charges that control the conductance of the transistors. However, the limitation of this solution is the relatively large and complex biasing circuit that is required to control the charges on the capacitor in addition to the non-linearity due to the variations of the drain-to-source voltage of the transistor. The recent work in \cite{jiang2019c3sram} employs both 8T-SRAM as a memory and cross-coupling capacitor as an accumulator to perform binary MAC operation using bitwise XNOR gate. To implement an analog MAC operation, this paper develops a novel cross-coupling capacitor (C3) computing, hence, named, the C3 processing unit (C3PU) coupled with a voltage-to-time converter (VTC) circuitry. The C3PU performs multiplication using capacitive coupling and accumulation through the transistor bitline in the array. The main contributions of this paper can be summarized as follows:

\begin{itemize}
	\item  According to the best of the authors' knowledge, this is the first circuit design that utilizes cross-coupling capacitor for IMC as both a memory and a computational element to perform analog MAC operation. 
	
	\item The proposed C3PU can be utilized in applications that heavily rely on vector-matrix multiplications, including but not limited to ANN, CNN, and DSP. The design is ideal for applications with fixed coefficients such as  pre-trained CNN weights and image compression \cite{halawani2018memristor}. 
	
	\item A 5.7$\mu$W low power voltage-to-time (VTC) converter is implemented at the input voltage terminal of the C3PU to generate a modulated pulse width signal. Such circuit guarantees a linear multiplication operation through CMOS transistor. 
	
	\item A 5$\times$4 crossbar architecture based on C3PU is designed and simulated in 65nm technology to employ 4 MAC units where each unit performs 5 multiplications and 4 additions. Simulation results show that the energy efficiency of the 5$\times$4 C3PU is 66.4 fJ/MAC at 0.3 V voltage supply with an error compared to computation in MATLAB of 5.7\%.

	\item The proposed C3PU usage has been demonstrated through iris flower classification on a two-layers ANN. The synaptic weights are trained offline and then mapped into capacitance ratio values for the inference phase. The ANN classifier circuit is designed and simulated in 65 nm CMOS technology. It achieves a high inference accuracy of 90\% compared to the baseline accuracy of 96.67\% obtained from MATLAB.

\end{itemize}
%

The rest of this paper is organized as follows. Section \ref{sec:C3PU} proposes the C3PU circuit design and explains how the MAC operation is performed. Section \ref{sec:crossbar} discusses the implementation of the MAC operations in a 5$\times$4 C3PU crossbar architecture. Section \ref{sec:app} shows an example of C3PU's potential application targeting iris flower classification using ANN architecture in 65nm technology. Finally, Section \ref{sec:conclusion} concludes the paper.

\section {Proposed C3PU Circuit and Operation}
\label{sec:C3PU}

The following subsections discuss the operational details of the proposed C3PU. The basic principle of the contribution is based on using a coupling capacitance to transfer the voltage to the transistor's gate. The generated voltage is linearly proportional to the current passed through the transistor. 

\subsection{C3PU Operation}

\begin{figure}[t]
	\centering
	\includegraphics[width=3.3in]{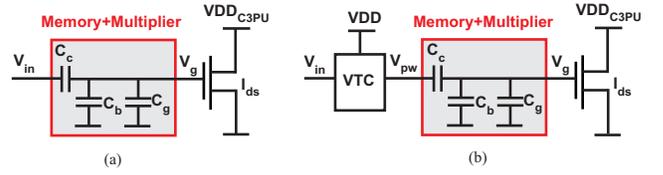}
	\caption{Proposed C3PU to perform a single multiplication operation by processing the input voltage $V_{in}$ in the (a) voltage domain and (b) time domain using VTC block.}
	\label{fig:C3PU_cell}
\end{figure}

Figure.~\ref{fig:C3PU_cell}a shows the proposed C3PU circuit that performs in-memory  multiplication operation. The C3PU consists of a CMOS transistor and a capacitive unit that includes a cross-coupling capacitor $C_{c}$, a capacitor $C_{b}$ connected between the gate of the transistor and the ground, and a transistor's gate capacitor $C_{g}$. The modulated input voltage amplitude $V_{in}$, which is the first multiplication operand, is applied at the terminal of the capacitive unit. The second operand is stored in the capacitive unit as an equivalent capacitance ratio $X_{eq}$=$\frac{C_{c}}{C_{c}+C_{b}+C_{g}}$. The capacitive computational unit multiplies the two operands and generates a voltage $V_{g}$ that is a function of $V_{in}$, $C_{c}$, $C_{b}$, and $C_{g}$ as given in Eq.~\ref{eq:dot_product}. $V_{g}$ is applied to the gate of CMOS transistor producing a drain-source current $I_{ds}$ as given in Eq.~\ref{eq:Ids} where $G_{m}$ is the transistor's transconductance. Note that $I_{ds}$ is proportional to the multiplication of its two operands $V_{in}$ and $X_{eq}$. Since the multiplication is linear, the transistor must also operate in a linear mode in order to transfer the multiplication correctly to the output in an electrical current form. 

\begin{equation}
\label{eq:dot_product}
V_{g}=V_{in}\frac{C_{c}}{C_{c}+C_{b}+C_{g}}
\end{equation}

\begin{equation}
\label{eq:Ids}
I_{ds}=G_{m}\times V_{g}= G_{m}\times V_{in}\frac{C_{c}}{C_{c}+C_{b}+C_{g}} 
\end{equation}

\begin{figure}[hbt]
	\centering
	\includegraphics[width=3.3in]{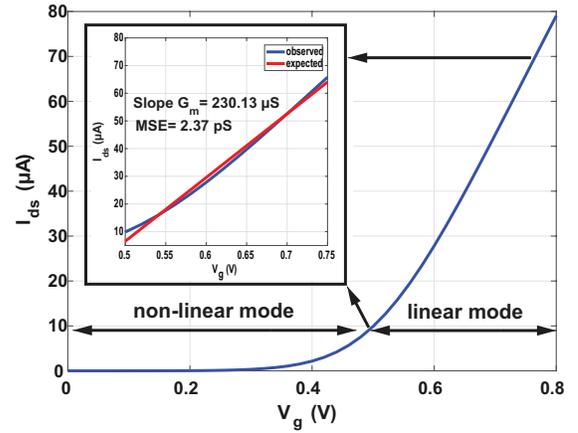}
	\caption{$I_{ds}$ versus $V_{g}$ at $VDD_{C3PU}$=0.3V. The transistor operates either in linear or non-linear mode based on the multiplication output of the two operands.}
	\label{fig:Ids_vs_Vg}
\end{figure}

The value of $V_{g}$ determines the operational mode of the transistor and affects its transconductance value and hence its linearity. Figure.~\ref{fig:Ids_vs_Vg} depicts the $I_{ds}$ of the transistor versus $V_{g}$ at $VDD_{C3PU}$=0.3 V. As shown in the figure, $I_{ds}$ is approximately linear only when $V_{g}$ is between 0.5 V and 0.8 V with a transconductance slope of 230.13 $\mu$S and a mean square error (MSE) of 2.37 pS between the observed and expected ones. The linearity over a small range of $V_{g}$ creates some design constraints. First, the input voltage has to be selected within a certain high value range. This means that $V_{in}$ requires normalization to tolerate the low $V_{in}$ values resulting in a mapping error. Second, even though $V_{in}$ is high, the capacitance ratio $X_{eq}$ should also be high enough to provide a large $V_{g}$ value to run the transistor in linear mode.

To overcome the former issues that significantly affect the functionality of the proposed C3PU multiplier, the analog input voltage will be processed in the time domain rather than the voltage domain. This is achieved using a voltage-to-time converter (VTC), as shown in Fig.~\ref{fig:C3PU_cell}b, by converting the amplitude of analog input $V_{in}$ into time delay to generate a modulated pulse width signal $V_{pw}$. This way, the voltage level of $V_{pw}$ is ensured to be high and having a value equal to the VTC's supply voltage $VDD$=1 V. Consequently, the transistor will always operate in linear mode giving that $X_{eq}$ is selected within a specific high range between 0.5 and 0.75 and $VDD_{C3PU}$ is low with a value of 0.3 V. If $X_{eq}>$0.75, then the value of $V_{g}$ will saturate. The resultant $I_{ds}$ becomes a function of $V_{pw}$ as shown in Eq.~\ref{eq:Ids_time} that is linearly proportional to the time delay. The proposed VTC circuit design, as discussed in section~\ref{sec:VTC} achieves high conversion linearity over a wide range of $V_{in}$. This guarantees that the C3PU performs a valid multiplication between $V_{in}$ and $X_{eq}$ by: $a$) providing a linear conversion from $V_{in}$ to $V_{pw}$, and $b$) running the transistor in linear mode. 
\begin{equation}
\label{eq:Ids_time}
I_{ds}=G_{m}\times V_{g}= G_{m}\times V_{pw}\frac{C_{c}}{C_{c}+C_{b}+C_{g}} 
\end{equation}

Presenting the data $V_{in}$ in the time domain has several advantages over the voltage domain, since both time and capacitance scale better with technology. In addition, it has less variations and provides better noise immunity compared to the voltage domain where the signal-to-noise ratio is degraded due to voltage scaling \cite{naraghi2009time}.

%

\subsection{Proposed Voltage-to-Time Converter (VTC)}
\label{sec:VTC}

\begin{figure}[t]
	\centering	\includegraphics[width=3.3in]{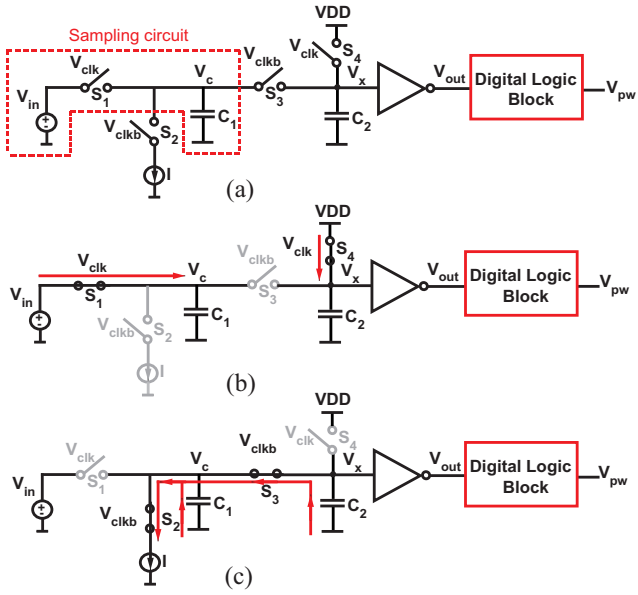}
	\caption{(a) Block diagram of the proposed VTC, (b) VTC's signal flow in the sampling phase, (c)VTC's signal flow in the evaluation phase.}
	\label{fig:top_level}
\end{figure}%

Figure.~\ref{fig:top_level} shows the block diagram of the proposed VTC circuit design. It consists of a sampling circuit, an inverter, and a current source.  To achieve voltage-to-time conversion, the VTC has two operating phases: sampling and evaluation. The basic principle is to transfer the charges from the input to the capacitor during the sampling phase and then discharge this capacitor through a current source during the evaluation phase. A simple inverter is used to transfer the time it takes to discharge the capacitor into a delay. The delay will be linearly proportional to the input voltage.

During the sampling phase, as shown in Fig.~\ref{fig:top_level}b, $S_{1}$ and $S_{4}$ turn on when the clock $V_{clk}$=1 V and $S_{2}$ and $S_{3}$ are off when the inverted clock $V_{clkb}$=0. The capacitor $C_{1}$ is precharged with a voltage $V_{c}$ that is equal to the input voltage value $V_{in}$. The capacitor $C_{2}$ is charged with a voltage $V_{x}$ that is equal to the supply voltage $VDD$. During the evaluation phase, as shown in Fig.~\ref{fig:top_level}c, $S_{1}$ and $S_{4}$ turn off when $V_{clk}$=0 and $S_{2}$ and $S_{3}$ turn on when $V_{clkb}$=1 V. The node $V_{c}$ is coupled to $V_{x}$. In this phase, the functionality of the VTC depends on $V_{in}$. When $V_{in}$ is high, i.e., $V_{in}$=$VDD$, then, $V_{c}$=$V_{x}$ and the initial charge across the capacitors is $Q_{i}$=$VDD(C_{1}+C_{2})$. On the other hand, when $V_{in}$ is small, i.e., $V_{in}$=0, the initial charge across the capacitors is $Q_{i}$=$V_{in}C_{1}$+$VDDC_{2}$. Due to the potential difference between $C_{1}$ and $C_{2}$, the charges are shared among them. Consequently, the current flows from $C_{2}$ to $C_{1}$ causing a voltage pump on $V_{c}$. Then, it starts discharging through the current source $I$ till it reaches the switching point of the inverter $V_{sp}$ resulting in a final charge $Q_{f}$=$V_{sp}(C_{1}+C_{2})$. After that, the inverter pulls up the delayed output voltage $V_{out}$. The time it takes to discharge $V_{x}$ to the inverter's switching point voltage is referred to as time delay $t_{d}$. This time delay, given in Eq.~\ref{eq:RC}, depends on four main parameters: voltage values of $VDD$ and $V_{in}$, voltage value of $V_{sp}$, capacitors' size of $C_{1}$ and $C_{2}$, and the average current $I_{avg}$ till it is discharged.  The $V_{sp}$ value is set by the aspect ratio of PMOS and NMOS transistors of the inverter ($\frac{\beta_{n}}{\beta_{p}}$) as given in Eq.~\ref{eq:Vth}. The $I_{avg}$ value depends on the amount of charges stored in the capacitors, which varies linearly with $V_{in}$ given that $VDD$ is fixed. Thus, $t_{d}$ has a linear relationship with $V_{in}$. 
\begin{equation}
t_{d}=\frac{Q_{i}-Q_{f}}{I_{avg}}= \frac{C_{1}V_{in}+C_{2}VDD-V_{sp}(C_{1}+C_{2})}{I_{avg}}
\label{eq:RC}
\end{equation}

\begin{equation}
V_{sp}=\frac{VDD-|V_{thp}|+\sqrt{\frac{\beta_{n}}{\beta_{p}}}V_{thn}}{1+\sqrt{\frac{\beta_{n}}{\beta_{p}}}}
\label{eq:Vth}
\end{equation}

\begin{figure}[t]
	\centering
	\begin{subfigure}[t]{0.5\textwidth}
		\centering
		\includegraphics[width=3.3in]{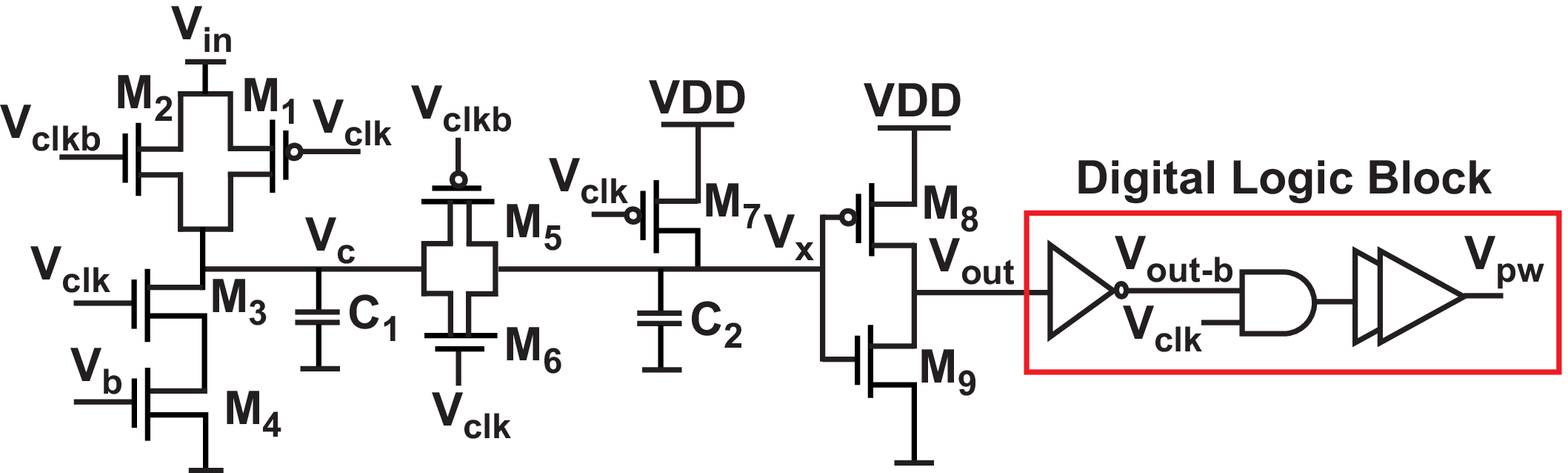}
		\caption{}
		\label{fig:VTC_cs}
	\end{subfigure}%
	\newline
	\begin{subfigure}[t]{0.5\textwidth}
		\centering
		\includegraphics[width=2.7in]{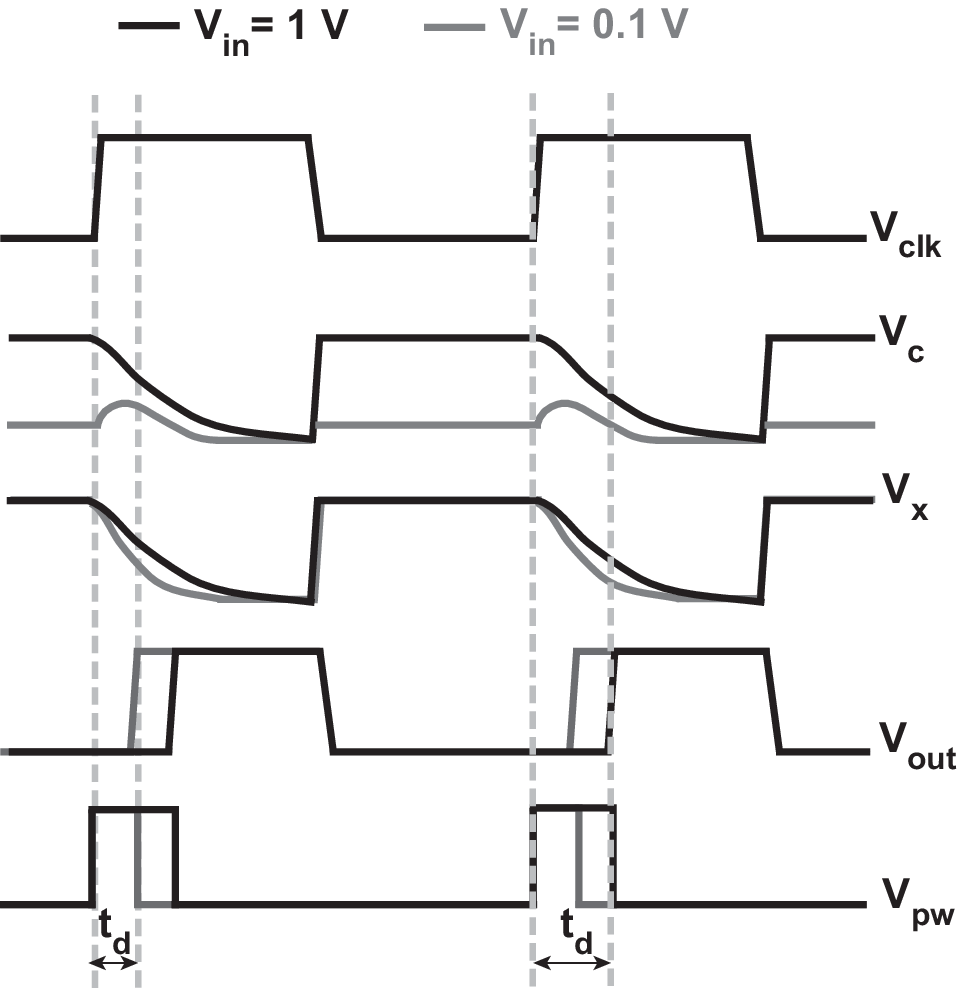}
		\caption{}
		\label{fig:VTC_waveform}
	\end{subfigure}%
	
	\caption{(a) Circuit design of the proposed VTC (b) input/output waveforms of VTC.}
	\label{fig:VTC_CS_two}
\end{figure}

%

To implement the proposed VTC using CMOS, Fig.~\ref{fig:VTC_cs} shows the detailed circuit diagram. The switches $S_{1}$ and $S_{3}$ are replaced by the pass gates ($M_{1}$, $M_{2}$) and ($M_{5}$, $M_{6}$), respectively. The switches $S_{2}$ and $S_{4}$ are replaced by $M_{3}$ and $M_{7}$, respectively. The current source is simply implemented using $M_{4}$ and controlled by a bias voltage $V_{b}$ to operate in the saturation region. The inverter is realized by  $M_{8}$ and $M_{9}$. To generate a pulse width signal $V_{pw}$, a digital logic block of inverter and AND gate is added. During the sampling phase, when $V_{clk}$=0 and $V_{clkb}$=1, $M_{3}$ is off, and $M_{7}$ is on, so that $C_{2}$ is charged to $VDD$. The pass gate ($M_{1}$, $M_{2}$) turns on to precharge $C_{1}$ with $V_{c}$=$V_{in}$. On the other hand, the pass gate ($M_{5}$, $M_{6}$) is off, which disconnects the node $V_{x}$ from $V_{c}$ to eliminate the short circuit current on the delay chain at low voltage levels of $V_{in}$. At this phase, $V_{x}$=$VDD$, which causes $V_{out}$=0. During the evaluation phase, when $V_{clk}$=1 and $V_{clkb}$=0, the pass gate ($M_{5}$, $M_{6}$) and $M_{3}$ turn on, whereas the pass gate ($M_{1}$, $M_{2}$) and $M_{7}$ turn off. In this phase, $V_{c}$ is coupled to $V_{x}$ and the charges redistribute between $C_{1}$ and $C_{2}$. Initially, if $V_{in}<VDD$, this means that $V_{c}<V_{x}$. As a result, a current flows from $C_{2}$ to $C_{1}$, making a charge pump on $V_{c}$ as shown in Fig.~\ref{fig:VTC_waveform} (see gray waveform when $V_{in}$=0.1 V). On the other hand, if $V_{in}$=$VDD$, then $V_{c}$ follows $V_{x}$ as shown in Fig.~\ref{fig:VTC_waveform} when $V_{in}$=1 V. In both cases, the capacitor current starts discharging through $M_{4}$, equating it with the drain-source current of $M_{4}$, $I_{ds4}$. This drops the value of $V_{x}$ till it reaches $V_{sp}$ of the inverter ($M_{8}$, $M_{9}$). Then, it pulls up $V_{out}$ that is connected to an inverter chain whose output $V_{out-b}$ is  ANDED with $V_{clk}$ to generate $V_{pw}$.  Figure.~\ref{fig:VTC_waveform} depicts the waveforms of the proposed VTC. Note that the proposed VTC controls the delayed $V_{out}$ at the rising edge of $V_{clk}$.

The proposed VTC circuit has been designed, implemented, and simulated in 65 nm industry-standard CMOS technology. The input voltage is set between 0 V to 1 V at $VDD$=1 V. Both capacitors $C_{1,2}$ and transistor $M_{4}$ sizes are selected to support a minimum time delay of 107 ps at the minimum $V_{in}$ of 0 V. A metal insulator metal (MIM) capacitors of $C_{1}$=27 fF and $C_{2}$=10 fF are utilized. The $M_{4}$ size of 500 nm/140 nm controlled by its gate voltage of $V_{b}$=0.5 V provides a current of 14 $\mu$A. The inverter is carefully sized to provide the desired $V_{sp}$. Hence, the aspect ratio of $M_{9}$ is 5$\times$ the aspect ratio of $M_{8}$ such that $V_{sp}$=0.35 V. Table~\ref{table:spec_vtc} summarizes the specifications of the proposed VTC design.

\begin{table}[hbt]
	\caption{Specifications of the proposed VTC.} 
	\centering 
\begin{tabular}{|c|c|}
	\hline
	$VDD$ (V)                                                                    & 1       \\ \hline
	$V_{in}$  (V)                                                                & 0-1     \\ \hline
	$C_{1}$ (fF)                                                                 & 27      \\ \hline
	$C_{2}$ (fF)                                                                 & 18      \\ \hline
	\begin{tabular}[c]{@{}c@{}}$W_{1,2,5,6}/L_{1,2,5,6}$ \\ (nm/nm)\end{tabular} & 600/60  \\ \hline
	$W_{3,7}/L_{3,7}$ (nm/nm)                                                    & 200/60  \\ \hline
	$W_{4}/L_{4}$ (nm/nm)                                                        & 500/140 \\ \hline
	$W_{8}/L_{8}$ (nm/nm)                                                        & 200/60  \\ \hline
	$W_{9}/L_{9}$ ($\mu$m/nm)                                                    & 1/60    \\ \hline
	$V_{b}$ (V)                                                                  & 0.5     \\ \hline
	$V_{sp}$ (V)                                                                 & 0.35    \\ \hline
\end{tabular}
	\label{table:spec_vtc} 
\end{table}

\begin{figure}[t!]
	    \centering
		\includegraphics[width=3.3in]{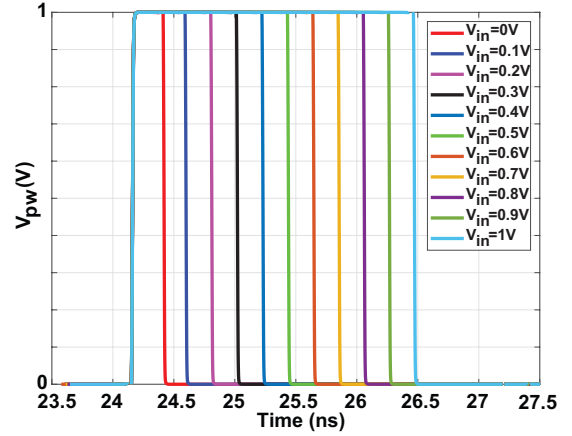}
		\caption{The modulated pulse width signal $V_{pw}$ for different $V_{in}$ values. The pulse width increases with the input voltage.}
		\label{fig:VTC_cs_Vpw}
\end{figure}

\begin{figure}[t!]
	\centering	\includegraphics[width=3.3in]{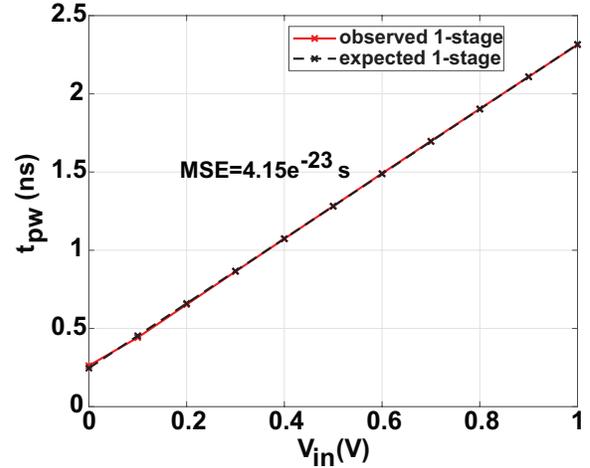}
	\caption{Observed (simulation) and expected (ideal) delay $t_{pw}$ vs $V_{in}$. The VTC shows high linearity with an MSE of 4.15e$^{-23}$.}
	\label{fig:MSE}
\end{figure}


Figure.~\ref{fig:VTC_cs_Vpw} depicts the modulated pulse width signal $V_{pw}$ at different $V_{in}$ values. As shown from the figure, the pulse width varies from 0.260 ns at $V_{in}$=0 V to 2.3 ns at $V_{in}$=1 V, resulting in a conversion gain of 2.05 ns/V. Figure.~\ref{fig:MSE} shows the output time delay $t_{pw}$ from the VTC versus the input voltage observed from the simulation in addition to the expected ones. As depicted from the figure, the time delay is linearly proportional to the input voltage. Note that the VTC is designed to operate in approximate computing architectures for AI applications that are statistical in nature and tolerable to variation and noise \cite{wang2019deep}\cite{reda2019approximate}. Noise simulation has been carried out to analyze the input-referred noise and the SNR of the VTC at Vin=1 V and frequency= 100 MHz. The input noise and signal power averages are obtained by integrating the noise and signal power spectrums over their frequency range. Spice simulation shows that the averaged input referred-noise and signal are 1.425 $\mu$$V^{2}$ and 5.67 $V^{2}$ resulting in an SNR value of 65.9 dB. The VTC has a low MSE value of 4.15e$^{-23}$ s, low power consumption of 5.7 $\mu$W, including the clock buffers and a small area of 0.0001 $mm^2$.

\begin{figure}[t!]
	\centering
		\includegraphics[width=3.3in]{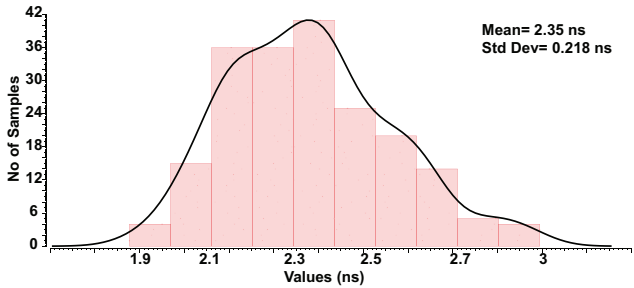}
		\caption{VTC's histogram mismatch variations at $V_{in}$=1 V. The x-axis represents the delay value and y-axis represents the number of samples. The total number of samples used is 200.}
		\label{fig:histogram_1v}
	
	\label{fig:Histogram}
\end{figure}

To quantify the impact of mismatch variation on the pulse width value, Monte Carlo Spice simulation is carried out with 200 samples. Figure.~\ref{fig:histogram_1v} shows the effect of mismatch variations on the time delay obtained from Monte Carlo simulation at $V_{in}$=1 V. As depicted from the figure, the standard deviation is low such as 0.218 ns from the mean of 2.358 ns at $V_{in}$=1 V. Hence, the ratio of the standard deviation to the mean is approximately 9\%. This variation can be reduced by cascading multiple stages of the VTC circuit as shown in Table~\ref{table:vtc_var}. As the number of the VTC stages increases, the variation decreases down to 4.4\% for 4-stages. For 4-stages VTC with 200 samples, 3-sigma variations of 13.2\% can be covered which is equivalent to 2-sigma variations for 2-stages VTC. 
Table~\ref{table:state_of_the_art} shows the comparison between the proposed design and prior works. Although the proposed VTC circuit has a lower conversion gain, the linearity range across $V_{in}$ is improved by 4$\times$ and 5.33$\times$ compared to \cite{yadav2020design} and \cite{mostafa2013highly}, respectively. Moreover, for IMC applications where the computation can be performed in a few ns, the pulse width of $V_{pw}$ doesn't need to be large, and so the conversion gain. The figure of merit (FoM) is developed for the VTC circuit and given in Eq.~\ref{eq:FoM}. It indicates accuracy of the VTC in providing conversion gain per power. The VTC's accuracy is 99.7\%, and hence the FoM equals 322 $\mu$s/V.W.
n t

\begin{equation}
FoM=accuracy\times\frac{Gain}{Power}
\label{eq:FoM}
\end{equation}

\begin{table}[hbt]
	\caption{Variations of the cascaded VTC stages for 200 samples.} 
	\centering 
	\begin{tabular}{|c|c|c|c|}
		\hline
		\textbf{\begin{tabular}[c]{@{}c@{}}VTC stage \\ number\end{tabular}} & \textbf{\begin{tabular}[c]{@{}c@{}}Mean \\ (ns)\end{tabular}} & \textbf{\begin{tabular}[c]{@{}c@{}}Standard \\ deviation (ns)\end{tabular}} & \textbf{\begin{tabular}[c]{@{}c@{}}Variation\\ (\%)\end{tabular}} \\ \hline
		1                                                                    & 2.04                                                          & 0.188                                                                       & 9.2                                                               \\ \hline
		2                                                                    & 4.06                                                          & 0.276                                                                       & 6.8                                                               \\ \hline
		3                                                                    & 6.058                                                         & 0.350                                                                       & 5.8                                                               \\ \hline
		4                                                                    & 7.98                                                          & 0.351                                                                       & 4.4                                                               \\ \hline
	\end{tabular}
	\label{table:vtc_var}
\end{table}

%

\begin{table}[t]
	\caption{Comparison between proposed and prior work.} 
	\centering 
\begin{tabular}{|c|c|c|c|c|}
	\hline
	Work                                                       & \cite{chen20203gs}                                     & \cite{yadav2020design}                               & \cite{mostafa2013highly}                                   & Proposed                                                    \\ \hline
	Technique                                                  & \begin{tabular}[c]{@{}c@{}}constant\\ slop\end{tabular} & \begin{tabular}[c]{@{}c@{}}super \\ MOS\end{tabular}  & \begin{tabular}[c]{@{}c@{}}starved \\ inverter\end{tabular} & \begin{tabular}[c]{@{}c@{}}sampling \\ circuit\end{tabular} \\ \hline
	\begin{tabular}[c]{@{}c@{}}Technology \\ (nm)\end{tabular} & 65                                                      & 45                                                    & 65                                                          & 65                                                          \\ \hline
	\begin{tabular}[c]{@{}c@{}}$VDD$\\ (V)\end{tabular}        & 1                                                       & 0.5                                                   & 1                                                           & 1                                                           \\ \hline
	\begin{tabular}[c]{@{}c@{}}$V_{in}$\\ (V)\end{tabular}     & 0-1                                                     & 0.1-0.5                                               & 0.2-0.35                                                    & 0-1                                                         \\ \hline
	\begin{tabular}[c]{@{}c@{}}Linearity \\ range\end{tabular} & \begin{tabular}[c]{@{}c@{}}high\\ 0-1\end{tabular}      & \begin{tabular}[c]{@{}c@{}}low\\ 0.2-0.4\end{tabular} & \begin{tabular}[c]{@{}c@{}}low\\ 0.2-0.35\end{tabular}      & \begin{tabular}[c]{@{}c@{}}high\\ 0-1\end{tabular}        \\ \hline
	\begin{tabular}[c]{@{}c@{}}Gain \\ (ns/V)\end{tabular}     & 0.144                                                   & 101.43                                                & 3.47                                                        & 2.05                                                        \\ \hline
	\begin{tabular}[c]{@{}c@{}}Power\\ ($\mu$W)\end{tabular}   & 8300                                                    & -                                                     & -                                                           & 5.7                                                         \\ \hline
	\begin{tabular}[c]{@{}c@{}}MSE\\ (s)\end{tabular}          & -                                                       & -                                                     & -                                                           & 4.15e$^{-23}$                                               \\ \hline
\end{tabular}
	\label{table:state_of_the_art} 
\end{table}


\begin{figure}[t!]
	\centering
	\includegraphics[width=3.3in]{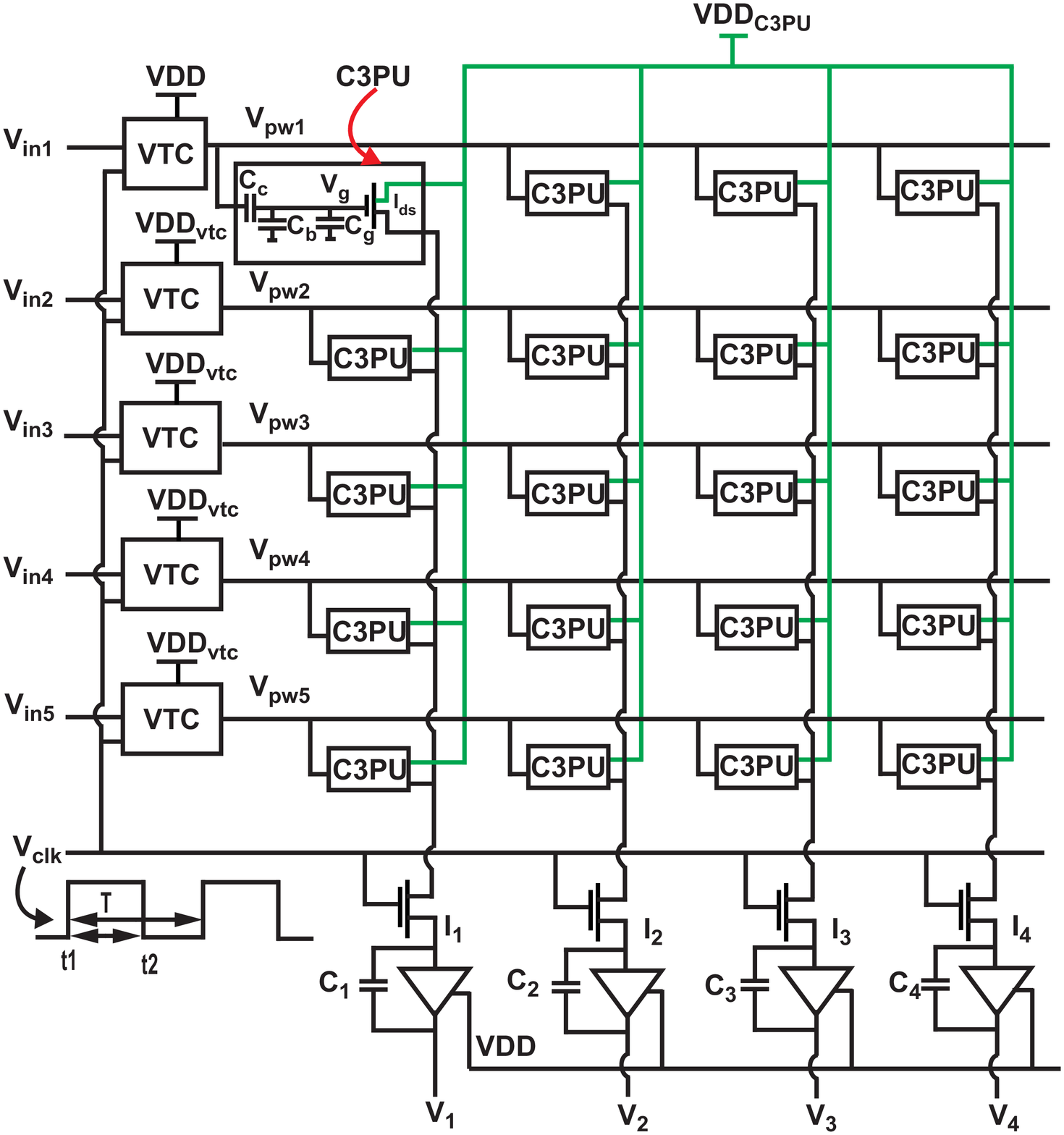}
	\caption{Proposed 5$\times$4 C3PU crossbar for MAC operations.}
\label{fig:C3PU_crossbar}
\end{figure}


\section{C3PU Crossbar Architecture for IMC Applications}
\label{sec:crossbar}

To demonstrate the advantage of the proposed design, a crossbar architecture of the C3PU and periphery circuit is designed. Computational crossbars naturally realize highly parallel vector-matrix operations and hence efficiently support high throughput with significant savings compared to the digital counterpart. This efficiency is achieved by performing the MAC operation in the same place where the data is stored. Therefore, the 5$\times$4 C3PU crossbar architecture is proposed, as shown in Fig.~\ref{fig:C3PU_crossbar}. The transistor source in each C3PU computational element is connected to the supply voltage $VDD_{C3PU}$. It is assumed that the analog input voltages $V_{in,1-5}$ come directly from the sensors. These inputs are converted into modulated pulse width signals $V_{pw,1-5}$ using 5 separate VTCs (discussed in ~\ref{sec:VTC}) instead of the need for the ADC as in the traditional design. The $V_{pw,1-5}$ represent the wordlines connected to the C3PU computational block to run it in linear mode. Each current produced by the C3PU is controlled by the multiplication of $V_{pw,i}$ and capacitance ratio $X_{eq,ij}$ ($i$ is the row and $j$ is the column) and then summed by the shared bitline. The resultant currents $I_{1-4}$ represent the complete MAC calculation of each column. The currents are integrated to generate an analog output $V_{1-4}$ to drive the actuator. Since the actuator function can be done in the analog domain, it reduces the overhead of going into the digital domain. 


The operation of the C3PU crossbar, given in Fig.~\ref{fig:C3PU_crossbar}, depends on two-phase functions: computation and isolation. In the computation phase, when the clock signal $V_{clk}$=1, the MAC operation is achieved by multiplying the  $V_{pw,i}$ pulse widths with the capacitance ratios $\frac{C_{c,ij}}{C_{c,ij}+C_{b,ij}+C_{g,ij}}$. Then, the transistors transfer this multiplication into a current that is summed on each bitline. 
The summed currents are integrated over a time $t_{1}-t_{2}$ using a virtual ground current integrator op-amp to provide the outputs as voltage levels $V_{1-4}$ as given in Eq.~\ref{eq:V_intg}. 
\begin{equation}
\label{eq:V_intg}
V_{j}=\frac{1}{C_{j}}\int_{t_{1}}^{t_{2}} I_{j} \, dt= \frac{1}{C_{j}}\int_{t_{1}}^{t_{2}} \sum_{i=1}^{i=5}I_{ds,ij} 
\end{equation}
The value of output voltages depends on two main parameters: $a$) time that the current will be accumulated $t_{1}-t_{2}$ and $b$) capacitor size $C_{j}$. The time $t_{1}-t_{2}$ is usually fixed and represents the pulse width of the clock. This time is set to be greater than the maximum pulse width of $V_{pw,i}$. The maximum pulse width of $V_{pw}$ is approximately 2 ns when the maximum input voltage $V_{in}$=1. Thus, the pulse width of the clock is set to 3 ns to ensure the completion of the computation and accumulation of the currents. In addition, the $C_{j}$ size plays an essential role in determining the scaling factor that is required to approximately allow $V_{1-4}$ to reach the expected output levels. The scaling factor is calculated by dividing the obtained MAC output voltages $V_{1-4}$ by the expected values, and hence the $C_{j}$ size is set. Once the approximate voltages are achieved, the C3PU elements are isolated from the outputs by setting $V_{clk}$=0 to enter the isolation phase. The isolation phase is essential to allow the proper functioning of the VTC and to initialize the output stage of the virtual ground op-amp. The period $T$, including computation and isolation time taken to operate the MAC calculations is 6 ns. Table~\ref{table:specs} shows the specifications of the C3PU crossbar architecture. The value of $C_{c}$ has a range between 2.5 fF and 8 fF, and the value of $C_{b}$ is fixed with 2.5 fF. Note that the proposed C3PU design targets hardwired fixed functions for AI applications where the weights are fixed. It can be modified to support applications that require programmable weights using emerging memcapacitor \cite{yawar2015investigation} \cite{salaoru2014coexistence}. However, this requires control circuits and a tunable voltage to program the capacitance value, which adds power overhead.


\begin{table}[t]
	\caption{5$\times$4 C3PU crossbar specifications.} 
	\centering 
\begin{tabular}{|c|c|}
	\hline
	$VDD_{C3PU}$ (V)                              & 0.3                         \\ \hline
	$VDD$ (V)                                     & 1                           \\ \hline
	$V_{in}$ (V)                                  & 0-1                         \\ \hline
	$V_{pw}$ (V)                                  & 1                           \\ \hline
	$t_{pw}$ (ns)                                 & 0-2.3                         \\ \hline
	$X_{eq}$                                      & 0.5-0.75                    \\ \hline
	$V_{g}$ (V)                                   & 0.5-0.75                    \\ \hline
	T (ns)                                        & 6                           \\ \hline
	\multicolumn{1}{|l|}{Transistor size (nm/nm)} & \multicolumn{1}{l|}{500/60} \\ \hline
\end{tabular}
	\label{table:specs} 
\end{table}

\begin{figure}[t!]
	\centering
	\includegraphics[width=3in]{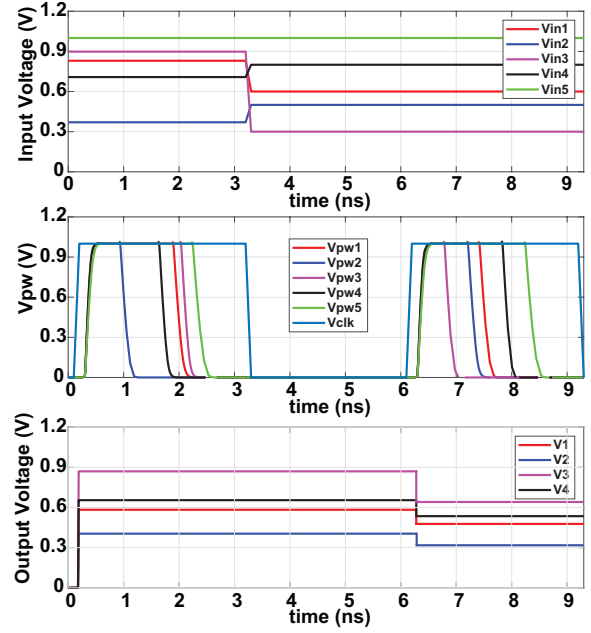}
	\caption{Input/output time domain signals of the 5$\times$4 C3PU crossbar for two different input sets.}
	\label{fig:time_domain}
\end{figure}

  \begin{table}[t]
	\caption{5$\times$4 C3PU crossbar output error matrix for different input combinations selected from the test set. The error is calculated by comparing the output from C3PU simulation with the output from MATLAB simulation.} 
	\centering 
	\begin{tabular}{|c|c|c|c|c|c|c|c|c|}
		\hline
		\multicolumn{5}{|c|}{\textbf{Input Voltage (V)}}                                                                                                                                 & \multicolumn{4}{c|}{\textbf{Error (\%)}}                                                                                           \\ \hline
		\multicolumn{1}{|l|}{$V_{in1}$} & \multicolumn{1}{l|}{$V_{in2}$} & \multicolumn{1}{l|}{$V_{in3}$} & \multicolumn{1}{l|}{$V_{in4}$} & \multicolumn{1}{l|}{$V_{in5}$} & \multicolumn{1}{l|}{$V_{1}$} & \multicolumn{1}{l|}{$V_{2}$} & \multicolumn{1}{l|}{$V_{3}$} & \multicolumn{1}{l|}{$V_{4}$} \\ \hline
		0.83                            & 0.37                           & 0.9                            & 0.71                            & 1                              & 6.5                          & 1.1                          & 1.6                          & 5.9                          \\ \hline
		0.2                             & 0.66                           & 0                              & 0                              & 1                              & 1.8                          & 5.2                          & 0.02                         & 3.5                          \\ \hline
		0.42                            & 0.3                            & 0.7                            & 0.75                           & 1                              & 0.8                          & 3.5                          & 2.1                          & 2.6                          \\ \hline
		0                               & 0.41                           & 0.5                            & 0                              & 1                              & 0.4                          & 1.3                          & 9.4                          & 2.5                          \\ \hline
		0.38                            & 0.37                           & 0.54                           & 0.5                            & 1                              & 3.8                          & 6.1                          & 7.3                          & 3.6                          \\ \hline
	\end{tabular}
	\label{table:error_matrix}
\end{table}


The 5$\times$4 C3PU crossbar shown in Fig.~\ref{fig:C3PU_crossbar} with the specifications in Table~\ref{table:specs} is designed and implemented in 65nm technology. The input voltages are fed to the C3PU crossbar for 30 consecutive clock cycles representing the 30 input sets. Each cycle has different sets of input voltage levels that are converted into modulated pulse width signals. Figure.~\ref{fig:time_domain} shows the input/output time domain waveform of the 5$\times$4 C3PU crossbar for two different input sets. The input voltages are validated at the negative edge clock, and the modulated pulse width signals are generated at the positive edge clock. The average computing error in the 5$\times$4 C3PU crossbar is 5.7\%. The error is calculated and averaged for 30 input samples by comparing the observed MAC output from simulation with the expected values. Table~\ref{table:error_matrix} demonstrates the error matrix of the C3PU outputs when compared to the expected ones from MATLAB simulation at different input combinations selected from the test set. The energy efficiency of the 5$\times$4 C3PU crossbar and the 5 VTC blocks is 26.3 fJ/MAC and 40.1 fJ/MAC, respectively, resulting in total energy efficiency of 66.4 fJ/MAC. Each MAC unit/column includes 5 multiplications and 4 additions. To further increase the number of operations, the crossbar array size can be enlarged. Some design constraints need to be considered when increasing the C3PU crossbar size. Adding more rows to the C3PU array increases the accumulated currents, which require a larger capacitor size in the integrator circuit to achieve the desired output voltage. For example, every additional 5 rows demand an additional 300fF capacitor. Therefore, there is a tradeoff between the number of rows and the integrator's capacitor size. Increasing the number of columns is also limited as the line resistance affects the driving signal of the $V_{pw}$. The resistance due to the line connected from the VTCs to the columns increases with the number of columns, and this degrades the pulse width of $V_{pw}$ signal. Simulation results show that the C3PU crossbar with 32 columns will suppress the pulse width of $V_{pw}$ by 10.8\%. The maximum number of columns that the C3PU crossbar can afford is 46 with degradation of 13.4\% in the pulse width. Another option to accommodate large MAC operations is to duplicate the C3PUs similar to memory arrays. For example, multiple C3PU arrays can be placed to increase the number of columns and rows where a repeater can be used instead of the VTC to generate the pulse width signal.


%


To compare the proposed 5$\times$4 C3PU crossbar, a 5$\times$4 fixed point (FXP) crossbar units have been implemented using ASIC design flow in 65 nm CMOS. Table~\ref{table:comparison} shows the 3$\times$3-bit, 4$\times$4-bit, 8$\times$4-bit, and  8$\times$8-bit FXP crossbars performance compared to the 5$\times$4 C3PU crossbar. The error of the FXP MAC unit is calculated by comparing the observed output from the RTL simulation for each column in the crossbar with the expected ones from MATLAB simulation. The resultant error values are then averaged over 30 input sets.The average error of the C3PU, 5.6\%, is comparable to the error percentage produced by the 8$\times$4-bit MAC unit, 6.52\%. Furthermore, the MSE values of the C3PU and 8$\times$4-bit MAC crossbars are almost equal with 0.082 and 0.099, respectively. The advantage of the C3PU is the lower energy and area consumption by 3.4$\times$ and 3.6$\times$, respectively, compared with the 8$\times$4-bit MAC unit. 

Table~\ref{table:prior_work} compares the prior and proposed work. The proposed C3PU utilizes an AMS circuit to perform analog MAC with two analog inputs, whereas the work in \cite{valavi2018mixed} and \cite{jiang2019c3sram} uses an AMS circuit to conduct binary MAC with 1-bit$\times$1-bit inputs. Comparing the C3PU with its equivalent digital baseline (8-bit$\times$4-bit) in terms of accuracy, the energy efficiency is improved by 3.4$\times$.

\begin{table}[t]
	\caption{Evaluation of 5$\times$4 FXP crossbar MAC units with different input and weight resolutions.} 
	\centering 
	\begin{tabular}{|c|c|c|c|c|}
		\hline
		\textbf{\begin{tabular}[c]{@{}c@{}}MAC Unit \\ Type\end{tabular}} & \textbf{\begin{tabular}[c]{@{}c@{}}Energy\\ (\small{fJ/MAC})\end{tabular}} & \textbf{\begin{tabular}[c]{@{}c@{}}Error \\ (\small{\%})\end{tabular}} & \textbf{MSE} & \textbf{\begin{tabular}[c]{@{}c@{}}Area\\ (\small{$\mu m^2$/MAC})\end{tabular}} \\ \hline
		\textbf{3$\times$3-bit}                                           & 60.9                                                               & 64.7                                                                      & 14.64        & 127.7                                                                   \\ \hline
		\textbf{4$\times$4-bit}                                           & 107                                                                & 10                                                                        & 0.24         & 246.2                                                                   \\ \hline
		\textbf{8$\times$4-bit}                                           & 226.2                                                              & 6.52                                                                      & 0.099        & 655.8                                                                   \\ \hline
		\textbf{8$\times$8-bit}                                           & 526                                                                & 0.74                                                                      & 0.002        & 1380.7                                                                  \\ \hline
		\textbf{C3PU}                                                     & 66.4                                                               & 5.7                                                                       & 0.082       & 180                                                                     \\ \hline
	\end{tabular}
	\label{table:comparison} 
\end{table}

\begin{table}[t]
	\caption{Comparison between the prior and proposed work.} 
	\centering 
	\begin{tabular}{|c|c|c|c|c|}
		\hline
		Work                                                                 & \cite{valavi2018mixed}                                                   & \cite{jiang2019c3sram}                                  & \multicolumn{2}{c|}{This Work}                                                                                                               \\ \hline
		\begin{tabular}[c]{@{}c@{}}Technology\\ (nm)\end{tabular}            & 65                                                                        & 65                                                       & \multicolumn{2}{c|}{65}                                                                                                                      \\ \hline
		\begin{tabular}[c]{@{}c@{}}MAC\\ Type\end{tabular}                   & Digital                                                                   & Digital                                                  & Analog                                                   & \begin{tabular}[c]{@{}c@{}}Digital\\ baseline\\ (8$\times$4-bit)\end{tabular} \\ \hline
		\begin{tabular}[c]{@{}c@{}}Operating \\ Voltage \\(V)\end{tabular}     & \begin{tabular}[c]{@{}c@{}}1.2\\ 0.94/0.68\end{tabular}                   & \begin{tabular}[c]{@{}c@{}}1\\ 0.8/0.6\end{tabular}      & \begin{tabular}[c]{@{}c@{}}1\\ 0.5/0.3\end{tabular}      & 1                                                                                 \\ \hline
		\begin{tabular}[c]{@{}c@{}}Energy \\Efficiency\\ (TOPS/W)\end{tabular} & 658                                                                       & 671                                                      & 136                                                      & 40                                                                                \\ \hline
		\begin{tabular}[c]{@{}c@{}} Cell \\Area\end{tabular}   & \begin{tabular}[c]{@{}c@{}}1.8$\times$ $>$\\ 6T\\SRAM\end{tabular} & \begin{tabular}[c]{@{}c@{}}2.9\\ $\mu$$m^2$\end{tabular} & \begin{tabular}[c]{@{}c@{}}4.1\\ $\mu$$m^2$\end{tabular} & \begin{tabular}[c]{@{}c@{}}132\\ $\mu$$m^2$\end{tabular}                          \\ \hline
	\end{tabular}
	\label{table:prior_work} 
\end{table}

\section{C3PU Demonstrator for ANN Applications}
\label{sec:app}

The advantage of the C3PU is demonstrated by accelerating the MAC operations found in an ANN using iris database \cite{murphy1994uci}. The data set consists of 150 samples divided equally between the three different classes of the iris flower, namely, Setosa, Versicolour, and Virginica. Each sample holds the following features all in cm: sepal length, sepal width, petal length, and petal width. The architecture of the ANN consists of two layers: four nodes for the input layer, each representing one of the input features, followed by three hidden neurons, and lastly, three output neurons for each class. To implement the MAC operations in the ANN, the iris features are considered as the first operands, which are mapped into voltage values, and the weights are considered as second operands that are stored as capacitance ratios in the capacitive unit of the C3PU. A simple linear mapping algorithm is used between the neural weights and capacitance ratios \cite{hu2016dot}.


The training phase is performed offline using MATLAB by dividing the data set between 80\% training, and 20\% testing. Post-training weights can have values with both positive and negative polarities. Hence, before mapping these weights into capacitance ratio values, they need to be shifted by the minimum weight value $w_{min}$. After performing the multiplication between the inputs and shifted weights, the effect of the shifting operation must be removed by subtracting the following term from all weights $\left|w_{min}\right|$$\times$$\sum\limits_{i=1}^n$ $IN$, where $IN$ is the input to the hidden/output layer and $n$ is the number of input/hidden nodes. Mapping such operation into C3PU architecture requires adding one column to the hidden and output crossbars to store the $w_{min}$ value in each layer. 

\begin{figure}[t!]
	\centering
	\includegraphics[width=3.3in]{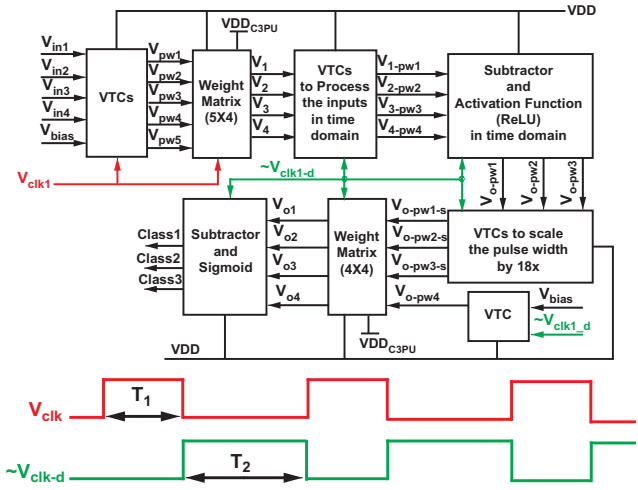}
	\caption{Algorithm flow of ANN classifier for iris flower data set illustrating the functional signals carried in the forward pass (inference) phase. 
	}
	\label{fig:classifier}
\end{figure}

Figure.~\ref{fig:classifier} depicts the algorithm flow of the ANN classifier for the iris data set. It has two operational phases: phase 1 and phase 2. In phase 1, when $V_{clk}$=1 and $\sim V_{clk-d}$=0, the inputs are processed in the first layer. In phase 2, when $V_{clk}$=0 and $\sim V_{clk-d}$=1, the outputs from the first layer are taken and processed in the second layer to generate the required output iris classes. In phase 1, the four input features are mapped into four voltage levels $V_{in1-4}$. These voltages are then converted into four pulse width modulated signals $V_{pw1-4}$ using four VTC blocks discussed in section~\ref{sec:VTC}. The bias voltage $V_{bias}$ is added as an input to better fit the ANN model, which is also converted into a pulse width modulated signal $V_{pw5}$. The $V_{pw1-5}$, first operands, are connected to the 5$\times$4 weight matrix C3PU as explained previously in Fig.~\ref{fig:C3PU_crossbar}. The weights, second operands, in this case, are stored as equivalent capacitance ratios $X_{eq}$ in the C3PU. The output voltages $V_{1-4}$ from the current integrator used at the end of each column in the C3PU weight matrix will act as inputs to the second layer. The current integrator inherently takes care of the scaling factor, which is decided depending on the factor between the shifted output values from a neural network and the output from the C3PU. This is important to compensate for the mapping between the values. 

Once $V_{1-4}$ are generated, the classifier switches to phase 2 to process them to the second layer. But before that, the impact of shift operation that is implemented on the weights needs to be removed by subtracting $V_{4}$ from $V_{1-3}$. Then, the subtracted outputs are passed through the ReLu activation function. In the proposed ANN classifier, the subtraction operation and ReLu function are implemented in the time domain. To achieve such implementation, $V_{1-4}$ are first converted to pulse width modulated signals using VTCs and then passed to the time domain subtractor and ReLu activation function to generate $V_{o-pw1-3}$. These output signals may have small pulse widths due to the subtraction operation which does not correspond to the expected subtraction outputs. Therefore, the pulse widths of the $V_{o-pw1-3}$ are scaled by a constant factor depending on the expected subtraction output from the ANN using MATLAB and the observed outcomes from the ANN using C3PU. After that, the scaled pulse width signals $V_{o-pw1-3-s}$ are fed to the 4$\times$4 C3PU weight matrix. The output voltages from the weight matrix $V_{o1-4}$ are passed to the subtractor and then the softmax function to generate the proper class based on the input features.    

\begin{figure}[t!]
	\centering
	\includegraphics[width=3.3in]{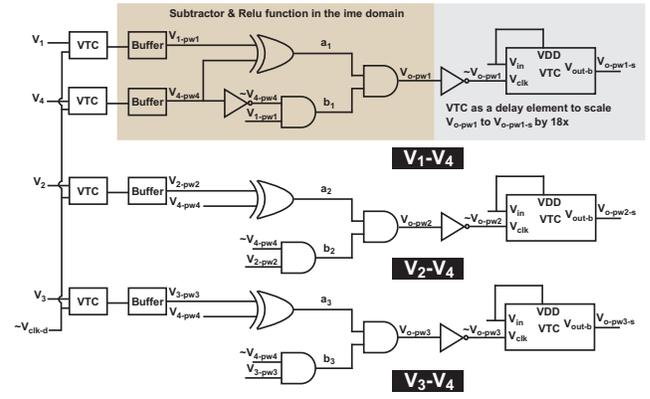}
	\caption{Circuit design of the time domain subtractor and ReLu function followed by the VTC as a delay element to increase the signals' pulse width by a factor of 18$\times$.}
	\label{fig:sub_relu_circuit}
\end{figure}

\begin{figure}[t!]
	\centering
	\includegraphics[width=3.3in]{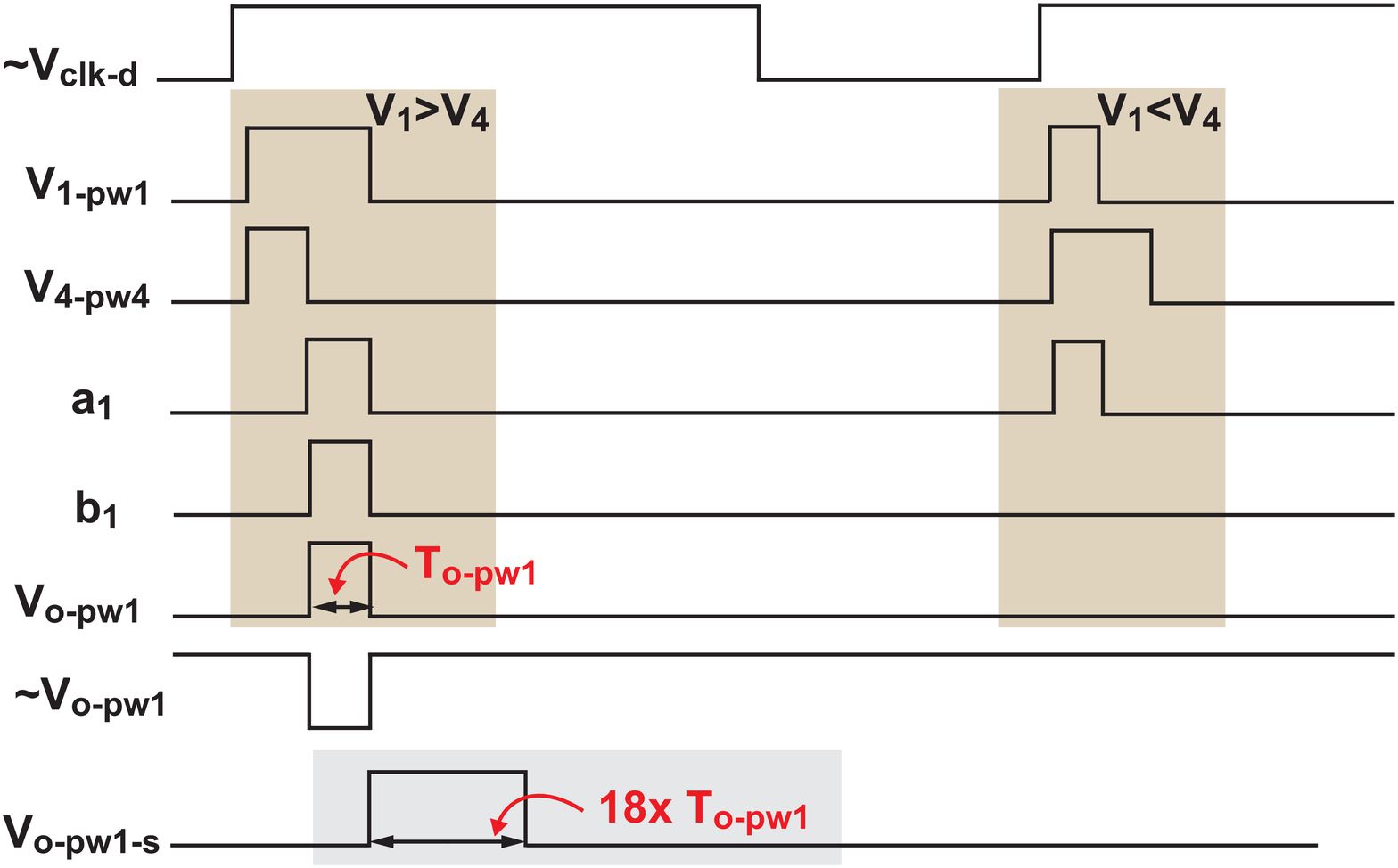}
	\caption{Waveform of the time domain subtractor and ReLu function when $V_{1}>V_{4}$ and $V_{1}<V_{4}$. When $V_{1}>V_{4}$, the pulse width of $V_{o-pw1}$ is generated whereas it is zero when When $V_{1}<V_{4}$. }
	\label{fig:sub_waveform}
\end{figure}   

Figure.~\ref{fig:sub_relu_circuit} shows the detailed circuit design implementation of the time domain subtractor, ReLu activation function, and delay element. Since $V_{4}$ is subtracted from three variables of $V_{1-3}$, then, each subtraction requires a separate digital circuit. The subtraction output can have a positive or a negative value. The ReLu activation function passes the positive value while assigning the negative value to zero. Such implementation is developed using AND, XOR, and inverter gates, as highlighted in the brown block in  Fig.~\ref{fig:sub_relu_circuit}. To detect the difference between the two pulse widths, the XOR gate is utilized and provides the subtraction output $a_{1-3}$. To determine the sign of the subtraction, $V_{4-pw4}$ is inverted and then ANDED with $V_{(1-3)-pw(1-3)}$ to generate a signal $b_{1-3}$. If any $b_{1-3}$=1, then the subtraction output is positive, whereas when $b_{1-3}$=0, the subtraction output is negative. Finally, AND gate is used to pass the positive subtraction output as $V_{o-pw1-3}$ while setting the negative subtraction output to zero. Figure~\ref{fig:sub_waveform} shows the output waveform example of the subtraction and ReLu function when $V_{1}>V_{4}$ and $V_{1}<V_{4}$. As depicted in the figure, when $V_{1}>V_{4}$, the modulated pulse width of $V_{1-pw1}$ is greater than the pulse width of $V_{4-pw-4}$. This means that the subtraction output is positive and passed with $V_{o-pw1}$=1 having a pulse width $T_{o-pw1}$ that represents the difference between the pulse width of $V_{1-pw1}$ and the pulse width of $V_{4-pw-4}$. On the other hand, when $V_{1}<V_{4}$, the subtraction difference is negative ($b_{1}$=0), resulting in $V_{o-pw1}$=0. 
Note that when the pulse width of the positive subtraction output is very narrow, it is rounded to zero, and the signal $V_{o-pw1}$ will disappear. This is referred to as quantization which is widely implemented in the digital domain to increase the computing energy efficiency while achieving an acceptable accuracy. The quantization in the time domain may affect the MAC outputs of the 2$^{nd}$ C3PU crossbar. However, since the computation is employed for AI applications, relative results are sufficient for the classification purpose.

After that, the pulse width $T_{o-pw1}$ of the signal $V_{o-pw1}$ is approximately scaled by a factor of 18$\times$ chosen based on the subtraction output values between the expected and observed ones. Such a large factor cannot be implemented using inverter delay. Consequently, a VTC circuit is utilized as a delay element to scale the pulse width of the $V_{o-pw1}$ by 18$\times$. To achieve such a scale, the capacitors’ values in the VTC are adjusted ($C_{1}$=50 fF and $C_{2}$=2 fF), and the input voltage is set to the supply voltage. The inverted subtraction output $\sim$$V_{o-pw1}$ is considered as the clock of the VTC. Depending on its pulse width value, the capacitors of $C_{1}$ and $C_{2}$ (as discussed in section \ref{sec:VTC}) are charged to a specific voltage level in the sampling phase. The higher the pulse width of the $\sim$$V_{o-pw1}$, the higher the voltage level across the capacitors and the longer time it takes to discharge through a current source in the evaluation phase. This means that the delay of the VTC's output $V_{o-pw1-s}$ is proportional to the pulse width of the $V_{o-pw1}$. The ANN classifier has been designed and simulated in 65 nm CMOS technology with a supply voltage of 1V except the 5$\times$4 and 4$\times$4 weight matrices that operate at a supply voltage of 0.3 V.  The input voltages $V_{in1-4}$ have a range of 0 V to 1 V in addition to $V_{bias}$=1 V. The five input voltages are converted into modulated pulse width signals $V_{pw1-5}$ that have pulse widths in the range of 165 ps to 2 ns. The modulated pulse width input signals $V_{o1-4}$ of the second weight matrix have a pulse width in the range of 1.6 ns to 7.5 ns. The pulse width $T_{1}$ of $V_{clk}$ is set to 3 ns, and the pulse width $T_{2}$ of $\sim$$V_{clk-d}$ is set to 9 ns. The proposed ANN classifier using C3PU shown in Fig.~\ref{fig:classifier} achieves an inference accuracy of 90\%, whereas the ideal implementation of the ANN classifier in MATLAB has an inference accuracy of 96.67\%. The variation of the supply voltage by 5\% affects the inference accuracy and reduces it by 3\%. Monte Carlo simulation has been carried out to study the mismatch variations on the inference accuracy. Although the MAC outputs' values from the C3PU crossbars have changed slightly, the inference accuracy remains 90\%. This is because the classification does not depend on the exact MAC outputs but rather on its relative values.

%
%

\section{Conclusion}
\label{sec:conclusion}
This paper presented an analog-mixed signal MAC unit using cross-coupling capacitor implementation named C3PU. The advantage of utilizing a cross-coupling capacitor for storage and processing element is that it can perform simultaneously as a high density and low energy storage. One operand in the C3PU is stored in the capacitive unit. While the second operand is a modulated pulse width signal using a voltage-to-time converter. The  multiplication outputs are transferred to an output current using CMOS transistors and then integrated using the current integrator op-amp. The 5$\times$4 C3PU was developed to run all data simultaneously, realizing fully parallel vector-matrix multiplication in one cycle. The energy consumption of the 5$\times$4 C3PU is 66.4 fJ/MAC at 0.3V voltage supply with an error of 5.7\% in 65nm technology. The inference accuracy for the ANN architecture has been evaluated using the proposed C3PU for an iris flower data set achieving a 90\% classification accuracy.

\bibliographystyle{IEEEtran}
\bibliography{TCAS1}

\begin{IEEEbiography}[{\includegraphics[width=1in,clip,keepaspectratio]{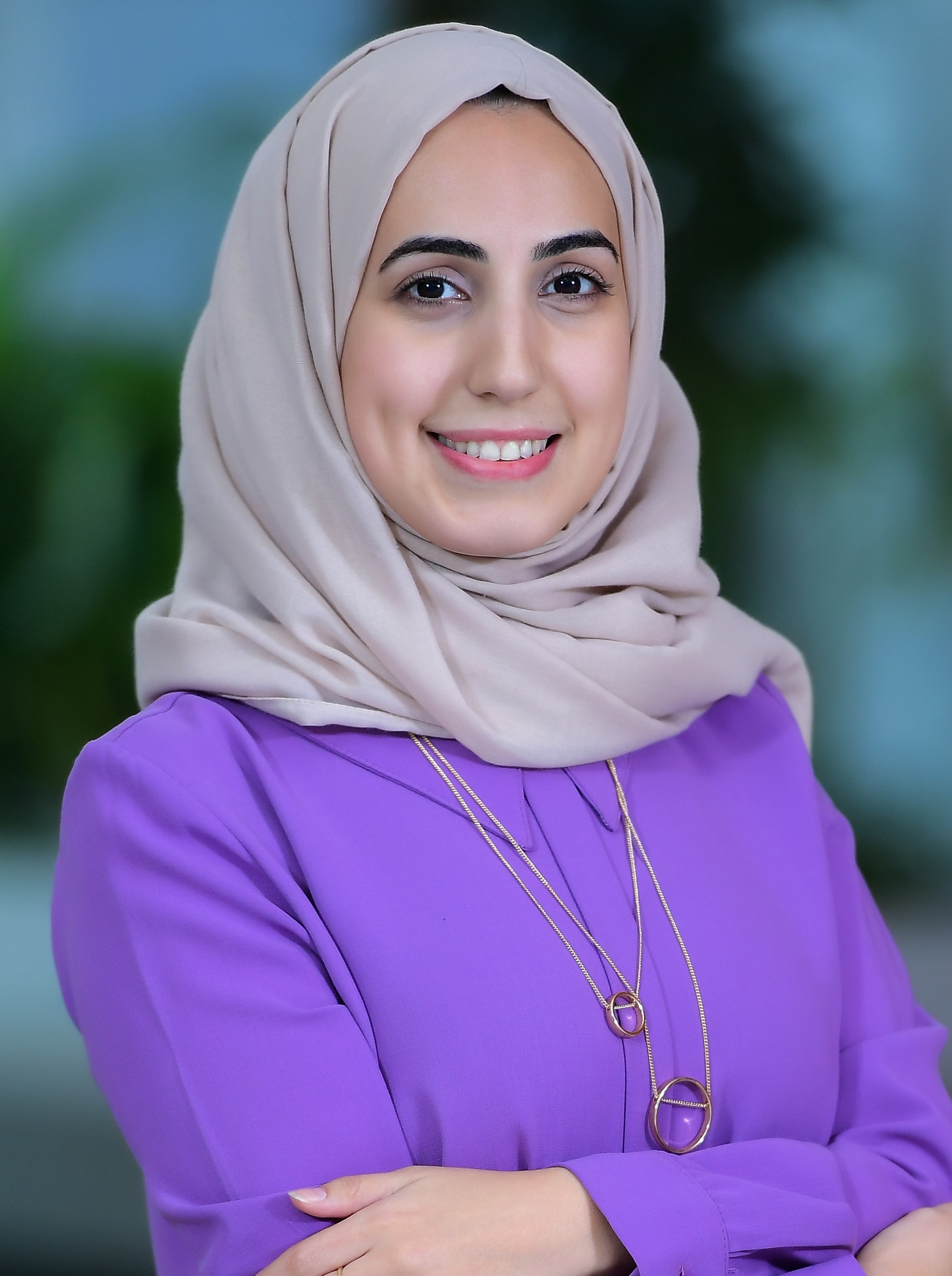}}]%
	{Dima Kilani (S’13-M’21)}
	received her PhD, M.S and B.S degrees in Electrical Engineering and Computer Science from Khalifa University, UAE, in 2019, 2015 and 2013, respectively. Her research focused on low-power mixed signal integrated circuit design including DC-DC power converters targeting high power efficiency. She is currently working as a postdoctoral fellow at the system-on-chip center (SoCC) in Khalifa University where she focuses on power management unit design for power-constrained devices. Dima works as a visiting scholar in Wayne State University, Detroit, MI researching system integration for wearable biomedical devices. She won the best paper session award in TECHCON-SRC in Texas, 2016.
\end{IEEEbiography}

\begin{IEEEbiography}[{\includegraphics[width=0.9in,clip,keepaspectratio]{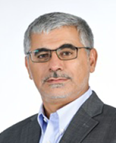}}]%
	{Baker Mohammad (SM'13)}
	earned his Ph.D. from University of Texas at Austin in 2008, his M.S. degree from Arizona State University, Tempe, and BS degree from the University of New Mexico, Albuquerque, all in ECE. He is an associate professor of electronic engineering at Khalifa University, and a consultant for Qualcomm Incorporated. Prior to joining Khalifa University, he was a Senior staff Engineer/Manager at Qualcomm and worked at Intel Corporation. He has over 16 year’s industrial experience in micro processor design with emphasis on memory, low power circuit, and physical design. His research interest includes power efficient computing, high yield embedded memory, emerging technology such as memristor, STTRAM, and computer architecture, energy harvesting and power management unit. 

\end{IEEEbiography}

\begin{IEEEbiographynophoto}{Yasmin Halawani} 
	(S’14-M’20) received her B.S. degree from the University of Sharjah, UAE, in 2012, the M.S. by Research degree in
	2014, and the Ph.D. degree in 2019, both from Khalifa University, UAE, and all in Electrical and Electronics Engineering. Her research projects focused on investigating the suitability of emerging memory technologies such as Memristor and STT-RAM for low-power applications. In addition, her research activities included the demonstration of the efficiency in-memory computing (IMC) for both analog and digital domains. She is currently working as a Post-Doctoral Fellow at Khalifa University in the area of memristor-based IMC architectures and artificial intelligence applications.
\end{IEEEbiographynophoto}

\begin{IEEEbiography}[{\includegraphics[width=1in,clip,keepaspectratio]{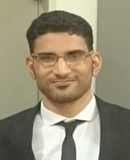}}]%
	{Mohammed F. Tolba}
 received his B.Sc. in 2014, from Electronics and communications engineering, Fayoum University and M.Sc. in 2018 from Micro-electronics System Design (MSD), Nile University. Currently, he is a research associate at SOC, Khalifa University. His research focused on digital design and implementation of deep learning, Convolution Neural Network (CNN), lightweight encryption, low-power approximation techniques, Graphics Processing Unit (GPU) architectures, computer arithmetic, fractional order circuits, Memristor, and chaotic circuits. Mohammed authored or co-authored over 28 journal and conference papers. Received the best paper award in Modern Circuits and Systems Technologies (MOCAST) 2017. In addition to the best master’s Thesis award July 2018. 
	
\end{IEEEbiography}

\begin{IEEEbiography}[{\includegraphics[width=1in,clip,keepaspectratio]{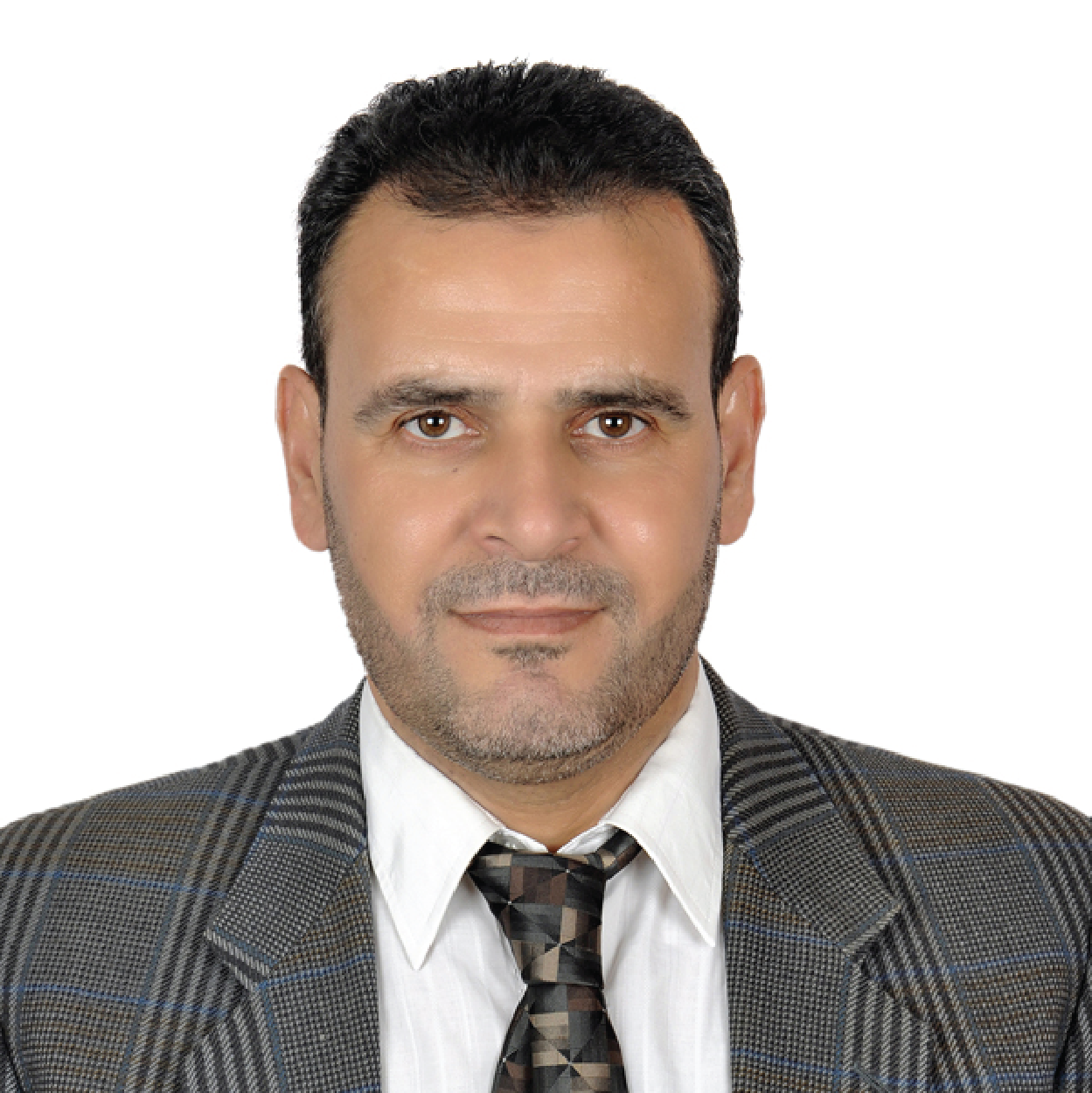}}]%
	{Hani Saleh (M'12)}
	is an assistant professor of electronic engineering at Khalifa University since 2012. Hani has a total of 19 years of industrial experience in ASIC chip design, microprocessor design, DSP core design, graphics core design and embedded system design. Prior to joining Khalifa University, he worked as a Senior Chip Designer (Technical Lead) at Apple incorporation and he worked for several leading semiconductor companies including Intel (ATOM mobile microprocessor design), AMD (Bobcat mobile microprocessor design), Qualcomm (QDSP DSP core design for mobile SOC’s) and Synopsys (a key member of Synopsys turnkey design group. Hani received a B.S in Electrical Engineering from the University of Jordan, a M.S in Electrical Engineering from the University of Texas at San Antonio, and a Ph.D. degree in Computer Engineering from the University of Texas at Austin. Hani research interest includes DSP algorithms design, DSP hardware design, computer architecture, computer arithmetic, SOC design, ASIC chip design, FPGA design and automatic computer recognition. 
\end{IEEEbiography}

\end{document}